\PassOptionsToPackage{usenames,dvipsnames,table,hyperref}{xcolor}
\documentclass[11pt,a4paper]{article}
\usepackage[hyperref]{acl2020}
\usepackage{times}
\usepackage{latexsym}

\usepackage{microtype}

\aclfinalcopy 

\usepackage{amsmath}
\usepackage{amssymb}
\usepackage{latexsym}
\usepackage{mathtools}
\usepackage{algorithm,algpseudocode}
\usepackage{textgreek}
\usepackage{bbm}
\usepackage{url}
\usepackage{xcolor}
\usepackage{tabularx}
\usepackage{multirow}
\usepackage{booktabs}
\usepackage{xspace}
\usepackage{kotex}
\usepackage[position=bottom]{subfig}
\usepackage[page]{appendix}
\usepackage{tabu}
\usepackage{color}
\usepackage{float}
\usepackage[normalem]{ulem} 
\usepackage{tikz}

\usepackage{amsfonts,dcolumn}

\newlength\myindent
\newcommand{\udensdash}[1]{%
    \tikz[baseline=(todotted.base)]{
        \node[inner sep=1pt,outer sep=0pt] (todotted) {#1};
        \draw[densely dashed] (todotted.south west) -- (todotted.south east);
    }%
}%

\newcolumntype{b}{X}
\newcolumntype{s}{>{\hsize=.5\hsize}X}

\usepackage{etoolbox}
\newtoggle{draft}
\togglefalse{draft}
\iftoggle{draft}{
\newcommand{\dk}[1]{\textcolor{Green}{[#1 \textsc{--DK}]}}
\newcommand{\taehee}[1]{\textcolor{BurntOrange}{[#1 \textsc{--Taehee}]}}
\newcommand{\thomas}[1]{\textcolor{Red}{[#1 \textsc{--Thomas}]}}
\newcommand{\lucas}[1]{\textcolor{Cyan}{[#1 \textsc{--Lucas}]}}
\newcommand{\hua}[1]{\textcolor{Purple}{[#1 \textsc{--Hua}]}}
}
{
\newcommand{\dk}[1]{}
\newcommand{\taehee}[1]{}
\newcommand{\thomas}[1]{}
\newcommand{\lucas}[1]{}
\newcommand{\hua}[1]{}
}

\newcommand{\method}{\textsc{P}os\textsc{C}al\xspace}
\newcommand{\Lagr}{\mathcal{L}}

\newcolumntype{P}[1]{>{\centering\arraybackslash}p{#1}}
\newcolumntype{M}[1]{>{\centering\arraybackslash}m{#1}}

\begin{document}

\title{Posterior Calibrated Training on Sentence Classification Tasks}

\author{
Taehee Jung$^1$ \quad Dongyeop Kang$^2$ \quad Hua Cheng$^3$ \quad Lucas Mentch$^1$ \quad Thomas Schaaf $^3$\\
$^1$Department of Statistics, University of Pittsburgh, Pittsburgh, PA, USA\\
$^2$School of Computer Science, Carnegie Mellon University, Pittsburgh, PA, USA \\
$^3$3M$|$M*Modal, Pittsburgh, PA, USA\\
\small{\texttt{$\{$taj41,lkm31$\}$@pitt.edu} \quad \texttt{ dongyeok@cs.cmu.edu}  \quad \texttt{ $\{$hcheng,tschaaf$\}$@mmm.com}}}

\date{}

\maketitle

\begin{abstract}
Most classification models work by first predicting a posterior probability distribution over all classes and then selecting that class with the largest estimated probability.
In many settings however, the quality of posterior probability itself (e.g., 65\% chance having diabetes), gives more reliable information than the final predicted class alone.
When these methods are shown to be poorly calibrated, most fixes to date have relied on posterior calibration, which rescales the predicted probabilities but often has little impact on final classifications.
Here we propose an end-to-end training procedure called posterior calibrated (\method) training that directly optimizes the objective while minimizing the difference between the predicted and empirical posterior probabilities. 
We show that \method not only helps reduce the calibration error but also improve task performance by penalizing drops in performance of both objectives.  
Our \method achieves about 2.5$\%$ of task performance gain and 16.1$\%$ of calibration error reduction on GLUE \cite{wang2018glue} compared to the baseline.
We achieved the comparable task performance with 13.2$\%$ calibration error reduction on xSLUE \cite{kang2019xslue}, but not outperforming the two-stage calibration baseline. 
\method training can be easily extendable to any types of classification tasks as a form of regularization term.
Also, \method has the advantage that it incrementally tracks needed statistics for the calibration objective during the training process, making efficient use of large training sets\footnote{Code is publicly available at \url{https://github.com/THEEJUNG/PosCal/}}.
\end{abstract}

\section{Introduction}\label{sec:intro}


Classification systems, from simple logistic regression to complex neural network, typically predict posterior probabilities over classes and decide the final class with the maximum probability. The model's performance is then evaluated by how accurate the predicted classes are with respect to out-of-sample,  ground-truth labels.
In some cases, however, the quality of posterior estimates themselves must be carefully considered as such estimates are often interpreted as a measure of confidence in the final prediction.
For instance, a well-predicted posterior can help assess the fairness of a recidivism prediction instrument~\cite{chouldechova2017fair} or select the optimal number of labels in a diagnosis code prediction~\cite{kavuluru2015empirical}. 

\lucas{I don't think this is a good example -- a better example might be predicting recidivism (whether someone will commit a crime if let out of prison) -- this topic has been frequently discussed with respect to the calibration issue in recent years.  See, e.g., ``Fair prediction with disparate impact:A study of bias in recidivism prediction instruments" by A. Chouldechova as a starting place...}
\thomas{I think having multiple short examples could be useful. Given that this is a submission for ACL, I suggest adding an example that is NLP/NLU related. "Classifying medical codes from medical notes is a prediction task which can be used to drive decisions in downstream applications. An applications can make different decisions based on the posterior probability of the prediction. These applications can benefit from well calibrated probabilities, allowing them to build their own (probabilistic and/or graphical) models for making decisions based on the predicted probabilities. For some models this means that if a well calibrated model is replaced with a better performing and also well calibrated model the downstream model does not need to change or tuned, and immediately benefits from the new model."}
\dk{+1 for adding multiple short examples from different fields (one from recidivism, another from medical). But, I like the Lucas's example with the reference. Also, then we need to replace the ``turn left'' example in the abstract accordingly.}
\taehee{Thomas also found the reference for medical code example. I am going to replace current example with two examples (recidivism, mdeical code).}

\citet{guo2017calibration} showed that a model with high classification accuracy does not guarantee good posterior estimation quality. 
In order to correct the poorly calibrated posterior probability, existing calibration methods \cite{zadrozny2001obtaining,platt1999probabilistic,guo2017calibration,kumar2019verified} generally rescale the posterior distribution predicted from the classifier after training.
Such post-processing calibration methods re-learn an appropriate 
distribution from a held-out validation set and then apply it to an unseen test set, causing a severe discrepancy in distributions across the data splits. 
The fixed split of the data sets makes the post-calibration very limited and static with respect to the classifier's performance.


We propose a simple but effective training technique called Posterior Calibrated (\method) training that optimizes the task objective while calibrating the posterior distribution in training. 
Unlike the post-processing calibration methods, \method directly penalizes the difference between the predicted and the true (empirical) posterior probabilities dynamically over the training steps.

\method is not a simple substitute of the post-processing calibration methods.
Our experiment shows that \method can not only reduce the calibration error but also increase the task performance on the classification benchmarks: 
compared to the baseline MLE (maximum likelihood estimation) training method, \method achieves 2.5\% performance improvements on GLUE \cite{wang2018glue} and 0.5\% on xSLUE \cite{kang2019xslue}, and at the same time 16.1\% posterior error reduction on GLUE and 13.2\% on xSLUE.


\section{Related Work}\label{sec:related}

Our work is primarily motivated by previous analyses of posterior calibration on modern neural networks.  \citet{guo2017calibration} pointed out that in some cases, as the classification performance of neural networks improves, its posterior output becomes poorly calibrated. 
There are a few attempts to investigate the effect of posterior calibration on natural language processing (NLP) tasks:
\citet{nguyen2015posterior} empirically tested how classifiers on NLP tasks (e.g., sequence tagging) are calibrated. 
For instance, compared to the Naive Bayes classifier, logistic regression outputs well-calibrated posteriors in sentiment classification task.
\citet{card2018importance} also mentioned the importance of calibration when generating a training corpus for NLP tasks. 

As noted above, numerous post-processing calibration techniques have been developed: 
traditional \textit{binning} methods \cite{zadrozny2001obtaining,zadrozny2002transforming} set up bins based on the predicted posterior $\hat{p}$, re-calculate calibrated posteriors $\hat{q}$ per each bin on a validation set, and then update every $\hat{p}$ with $\hat{q}$ if $\hat{p}$ falls into the certain bin.
On the other hand, \textit{scaling} methods \cite{platt1999probabilistic,guo2017calibration,kull2019beyond} re-scale the predicted posterior $\hat{p}$ from the softmax layer trained on a validation set.
Recently, \citet{kumar2019verified} pointed out that such re-scaling methods do not actually produce well-calibrated probabilities as reported since the true posterior probability distribution can not be captured with the often low number of samples in the validation set\footnote{\S\ref{sec:experiment} shows that the effectiveness of re-calibration decreases when the size of the validation set is small.} 
\lucas{This sentence is unclear}. 
To address the issue, the authors proposed a scaling-binning calibrator, but still rely on the validation set.


In a broad sense, our end-to-end training with the calibration reduction loss can be seen as sort of regularization designed to mitigate over-fitting. 
Just as classical explicit regularization techniques such as the lasso \cite{tibshirani1996regression} penalize models large weights, here we penalize models with posterior outputs that differ substantially from the estimated true posterior. 
\section{Posterior Calibrated Training}\label{sec:method}

In general, most of existing classification models are designed to maximize the likelihood estimates (MLE).
\lucas{should probably add a little bit here -- what is this sentence intended to convey?  It seems a bit crude to summarize all of classification in one sentence.}
Its objective is then to minimize the cross-entropy (Xent) loss between the predicted probability and the true probability over $k$ different classes.

During training time, \method minimizes the cross-entropy as well as the calibration error as a multi-task setup. 
While the former is a task-specific objective, the latter is a \textit{statistical objective} to make the model to be statistically well-calibrated from its data distribution.
Such data-oriented calibration makes the task-oriented model more reliable in terms of its data distribution.
Compared to the prior post-calibration methods with a fixed (and often small) validation set, \method \textit{dynamically} estimates the required statistics for calibration from the train set during training iterations.


Given a training set $\mathcal{D} =$ $\{(x_{1},y_{1})$$..$$(x_{n},y_{n})\}$ where $x_{i}$ is a p-dimensional vector of input features and $y_{i}$ is a k-dimensional one-hot vector corresponding to its true label (with $k$ classes), our training minimizes the following loss: 
\begin{equation}\label{eq:poscal}
    \Lagr_{\method} = \Lagr_{xent} + \lambda \Lagr_{cal}
\end{equation}
where $\Lagr_{xent}$ is the cross-entropy loss for task objective (i.e., classification) and $\Lagr_{cal}$ is the calibration loss on the cross-validation set. 
$\lambda$ is a weighting value for a calibration loss $\Lagr_{cal}$. 
In practice, the optimal value of $\lambda$ can be chosen via cross-validation.
More details are given in \S\ref{sec:experiment}.

Each loss term can be then calculated as follows:
\begin{align}\label{eq:poscal_details}
    \Lagr_{xent} &= -\sum_{i=1}^{n} \sum_{j=1}^{k} y^{(j)}_{i} log(\hat{p}^{(j)}_{i}) \\
    \Lagr_{cal} &= \sum_{i=1}^{n} \sum_{j=1}^{k} d(\hat{p}^{(j)}_{i},q^{(j)}_{i})
\end{align}
where $\Lagr_{xent}$ is a typical cross-entropy loss with $\hat{p}$ as an updated predicted probability while training.
$\Lagr_{cal}$ is our proposed loss for minimizing the calibration loss: 
$q$ is an true (empirical) probability and $d$ is an function to measure the difference (e.g., mean squared error or Kullback-Leibler divergence) between the updated $\hat{p}$ and true posterior $q$ probabilities.
The empirical probability $q$ can be calculated by measuring the ratio of true labels per each bin split by the predicted posterior $\hat{p}$ from each update. 
We sum up the losses from every class $j \in \{1,2..k\}$.



\begin{algorithm}[h]
 \small
\caption{{\small{Posterior Calibrated Training}}}\label{euclid}
\hspace*{\algorithmicindent} \textbf{Inputs} : \\
\begin{tabularx}{0.8\textwidth}{c X}
& \quad Train set $\mathcal{D}$, Bin $B$, Number of Classes $K$ \\
& \quad Number of epochs $e$, Learning rate $\eta$ \\
& \quad Number of updating empirical probabilities $u$ \\
\end{tabularx}
\\
\hspace*{\algorithmicindent} \textbf{Output} $\Theta$: Model Parameters
\begin{algorithmic}[1]
\State {Let $\mathcal{Q}$ : Empirical Probability Matrix $\in \mathbb{R}^{B \times K}$}
\State {Random initialization of $\Theta$}
\For{$i \in \{1,2,3,...e\}$}{}
    \State{Break $\mathcal{D}$ into random mini-batches b}
    \State{Find a set of steps $\mathcal{S}$ for updating $\mathcal{Q}$ by dividing total number of steps into $u$ equal parts}
    \For{b from $\mathcal{D}$}{}
        \State{$\Theta \leftarrow  \Theta - \eta \nabla_{\Theta} \mathcal{L}_{\method}(\Theta,\mathcal{Q})$}  
        \If{current step $\in \mathcal{S}$}
            \State{$\hat{p}$ = softmax($\Theta,\mathcal{D}$)}
            \State{$\mathcal{Q} \leftarrow CalEmpProb(\hat{p},B)$}
        \EndIf
    \EndFor
\EndFor
\end{algorithmic}
\label{alg:alg}
\end{algorithm}

We show a detailed training procedure of \method in Algorithm~\ref{alg:alg}.
While training, we update the model parameters (i.e., weight matrices in the classifier) as well as the empirical posterior probabilities by calculating the predicted posterior with the recently updated parameters.
For $\mathcal{Q}$, we exactly calculate 
a label frequency per bin $B$.
Since it is time-consuming to update $\mathcal{Q}$ at every step, we set up the number of $\mathcal{Q}$ updates per each epoch so as to only update $\mathcal{Q}$ at each batch. 

\section{Experiment}\label{sec:experiment}
We investigate how our end-to-end calibration training produces better calibrated posterior estimates without sacrificing task performance.


\textbf{Task: NLP classification benchmarks.}
We test our models on two different benchmarks on NLP classification tasks: GLUE~\cite{wang2018glue} and xSLUE \cite{kang2019xslue}. 
GLUE contains different types of general-purpose natural language understanding tasks such as question-answering, sentiment analysis and text entailment. 
Since true labels on the test set are not given from the GLUE benchmark, we use the validation set as the test set, and randomly sample 1\% of train set as a validation set. 
xSLUE \cite{kang2019xslue} is yet another classification benchmark but on different types of styles such as a level of humor, formality and even demographics of authors.
For the details of each dataset, refer to the original papers.

\textbf{Metrics.}
In order to measure the task performance, we use different evaluation metrics for each task. 
For GLUE tasks, we report \texttt{F1} for MRPC, \texttt{Matthews} correlation for CoLA, and \texttt{accuracy} for other tasks followed by \citet{wang2018glue}.
For xSLUE, we use \texttt{F1} score.

To measure the calibration error, we follow the metric used in the previous work~\citep{guo2017calibration}; Expected Calibration Error (\texttt{ECE}) by measuring how the predicted posterior probability is different from the empirical posterior probability: $\texttt{ECE} = \frac{1}{K}\sum_{k=1}^{K} \sum_{b=1}^{B} \frac{|B_{kb}|}{n} |q_{kb} - \hat{p}_{kb}|$,
where $\hat{p_{kb}}$ is an averaged predicted posterior probability for label $k$ in bin $b$, $q_{kb}$ is a calculated empirical probability for label $k$ in bin $b$, $B_{kb}$ is a size of bin $b$ in label $k$, and $n$ is a total sample size.
The lower \texttt{ECE}, the better the calibration quality.

\textbf{Models.}
We train the classifiers with three different training methods: \textbf{MLE}, \textbf{L1}, and \textbf{\method}. 
\textbf{MLE} is a basic maximum likelihood estimation training by minimizing the cross-entropy loss, \textbf{L1} is MLE training with L$_{1}$ regularizer, and \textbf{\method} is our proposed training by minimizing $\Lagr_{\method}$ (Eq \ref{eq:poscal}).
For \method training, we use Kullback-Leibler divergence to measure $\Lagr_{cal}$.
We also report \texttt{ECE} with a temperature scaling~\cite{guo2017calibration} (\textbf{tScal}), which is considered the state-of-the-art post-calibration method.

For our classifiers, we fine-tuned the pre-trained BERT classifier \cite{devlin2018bert}. Details on the hyper-parameters used are given in Appendix \ref{sec:appendix_hyper}.

\begin{table}[t!]
\centering
\small
\begin{tabularx}{\linewidth}{@{} r @{\hskip 0.3cm} c@{\hskip 0.1cm}c@{\hskip 0.1cm}c @{\hskip 0.4cm} c@{\hskip 0.1cm}c@{\hskip 0.1cm}c@{\hskip 0.1cm}c@{}}
\toprule
& \multicolumn{3}{c}{\scriptsize\texttt{Task Perf.} ($\uparrow$)} & \multicolumn{4}{c}{\scriptsize\texttt{Calib. ECE} ($\downarrow$)} \\
\cmidrule(rr){2-4} \cmidrule(lr){5-8}
Dataset &  \textbf{MLE} & \textbf{L$_1$}& \textbf{\method} &  \textbf{MLE} & \textbf{L$_1$}& \textbf{tScal} & \textbf{\method}\\
\midrule
\texttt{CoLA}   & 56.7& 55.3 &\textbf{58.0}   & .242&\textbf{.234}&.565&.231\\
\texttt{SST-2}  & 92.1&91.4&\textbf{92.4} & .144&.155&.143&\textbf{.106}\\
\midrule
\texttt{MRPC}& 88.2&88.2 &\textbf{88.9}  &.228&.229&.400&\textbf{.177} \\
\texttt{QQP}& 88.8&88.9&\textbf{89.1}  &.121&.122&\textbf{.054}&.107\\
\midrule
\texttt{MNLI}&\textbf{84.0}& 83.7&83.5 &.158&.160&\textbf{.080}&.165\\
\texttt{MNLI}$_{mm}$ &83.7&84.0&\textbf{84.2}    &.153&.153&\textbf{.062}&.149\\
\texttt{QNLI}&89.9&89.7&\textbf{90.0} &.138&.124&.159&.176\\
\texttt{RTE} & 61.7&62.4&\textbf{62.8} &.422&.441&\textbf{.175}&.394\\
\texttt{WNLI} &38.0&38.0&\textbf{56.9} &.287&.287&.269&\textbf{.083}\\
\midrule
\textbf{total}&75.9&75.6&\textbf{78.4} &.210&.212&.252&\textbf{.176} \\
\bottomrule
\end{tabularx}
\caption{\label{tab:result_glue} Task performance (left; higher better) and calibration error (right; lower better) on GLUE. 
We do not include STS-B; a regression task.
Note that \textbf{tScal} is only applicable for calibration reduction, because the post-calibration does not change the task performance, while \textbf{\method} can do both.
} 
\end{table}

\begin{table}[t!]
\centering
\small
\begin{tabularx}{\linewidth}{@{} r @{\hskip 0.1cm} c@{\hskip 0.05cm} c@{\hskip 0.05cm} c@{\hskip 0.1cm} c@{\hskip 0.05cm}c@{\hskip 0.05cm}c@{\hskip 0.05cm}c@{}}
\toprule
& \multicolumn{3}{c}{\scriptsize\texttt{Task Perf.}($\uparrow$)} & \multicolumn{4}{c}{\scriptsize\texttt{Calib. ECE}($\downarrow$)} \\
\cmidrule(lr){2-4} \cmidrule(lr){5-8}
Dataset &  \textbf{MLE} & \textbf{L$_1$} & \textbf{\method} &  \textbf{MLE} & \textbf{L$_1$} & \textbf{tScal}& \textbf{\method}\\
\midrule
\texttt{GYAFC} & 89.1&89.4& \textbf{89.5}        &.178&.170&.783&\textbf{.118}\\
\midrule
\texttt{SPolite}& 68.7&70.0& \textbf{70.9}  &.451&.431&\textbf{.133}&.238 \\

\midrule
\texttt{SHumor}&97.4 &\textbf{97.6}&\textbf{97.6}    &.050&.047&\textbf{.037}&.044\\
\texttt{SJoke}	&\textbf{98.4}&98.1&98.3   &.032&.037&\textbf{.019}&.029\\
\midrule
\texttt{SarcGhosh} & 42.5& 42.5& \textbf{42.6}   &.912&.912&\textbf{.898}&.910\\
\texttt{SARC} &71.3 &71.5 & \textbf{71.4}       &.372&.375&\textbf{.079}&.186\\
\texttt{SARC}{\_pol}& 72.7&72.8 &\textbf{73.8} &.434&.435&\textbf{.070}&.383\\
\midrule
\texttt{VUA}& 80.9&80.8&\textbf{81.4}          &.268&.276&.687&\textbf{.238}\\
\texttt{TroFi}&76.7&\textbf{78.8}&77.4           &.278&\textbf{.239}&.345&.265\\
\midrule
\texttt{CrowdFlower}&22.0&\textbf{22.7}&22.6&.404&.413&\textbf{.261}&.418\\
\texttt{DailyDialog}&48.3&47.8&\textbf{48.7}   &.225&.227&\textbf{.117}&.222\\
\midrule
\texttt{HateOffens}&93.0&\textbf{93.6}&93.5 
&.064&.059&.100&\textbf{.055}\\
\midrule
\texttt{SRomance}& 99.0&99.0&\textbf{100.0}  &.020&.020&.023&\textbf{.010}\\
\midrule
\texttt{SentiBank}& 96.7&\textbf{97.0}&96.6      &.061&.057&\textbf{.037}&.054\\
\midrule
\texttt{PASTEL}{\_gender} & 47.9 &\textbf{48.1}& 47.9    &.336&.305&.185&\textbf{.143}\\
\texttt{PASTEL}{\_age}	  &\textbf{23.5}&23.4& 22.9    &.354&.365&\textbf{.222}&.369\\
\texttt{PASTEL}{\_count} &56.1&56.6 &\textbf{58.3}&.054&.055&\textbf{.019}&.046\\
\texttt{PASTEL}{\_polit}& 46.6&\textbf{47.0}&46.8     &.394&.379&\textbf{.160}&.413\\
\texttt{PASTEL}{\_educ} &24.4&\textbf{25.2}& 24.7   &.314&.332&\textbf{.209}&.323\\
\texttt{PASTEL}{\_ethn} &\textbf{25.3}&24.8&24.8    &.245&.243&\textbf{.163}&.250\\
\midrule
\textbf{total}&64.0&64.3&\textbf{64.5}                &.272&.269&\textbf{.227}&.236\\
\bottomrule
\end{tabularx}
\caption{\label{tab:result_xslue} 
Task performance (left; higher better) and calibration error (\texttt{ECE}; lower better) on xSLUE. 
We do not include EmoBank; a regression task.
} 
\end{table}

\textbf{Results.}
Table \ref{tab:result_glue} and \ref{tab:result_xslue} show task performance and calibration error on two benchmarks: GLUE and xSLUE, respectively.
In general, \method outperforms the MLE training and MLE with L$_1$ regularization in GLUE for both task performance and calibration, though not in xSLUE. 
Compared to the tScal, \method shows a stable improvement over different tasks on calibration reduction, while tScal sometimes produces a poorly calibrated result (e.g., CoLA, MRPC).

\begin{figure}[t!]
\vspace{-3mm}
\center
{
\subfloat[Predictions in RTE]{
\includegraphics[trim=6mm 6mm 10mm 14mm,clip,width=.48\linewidth]{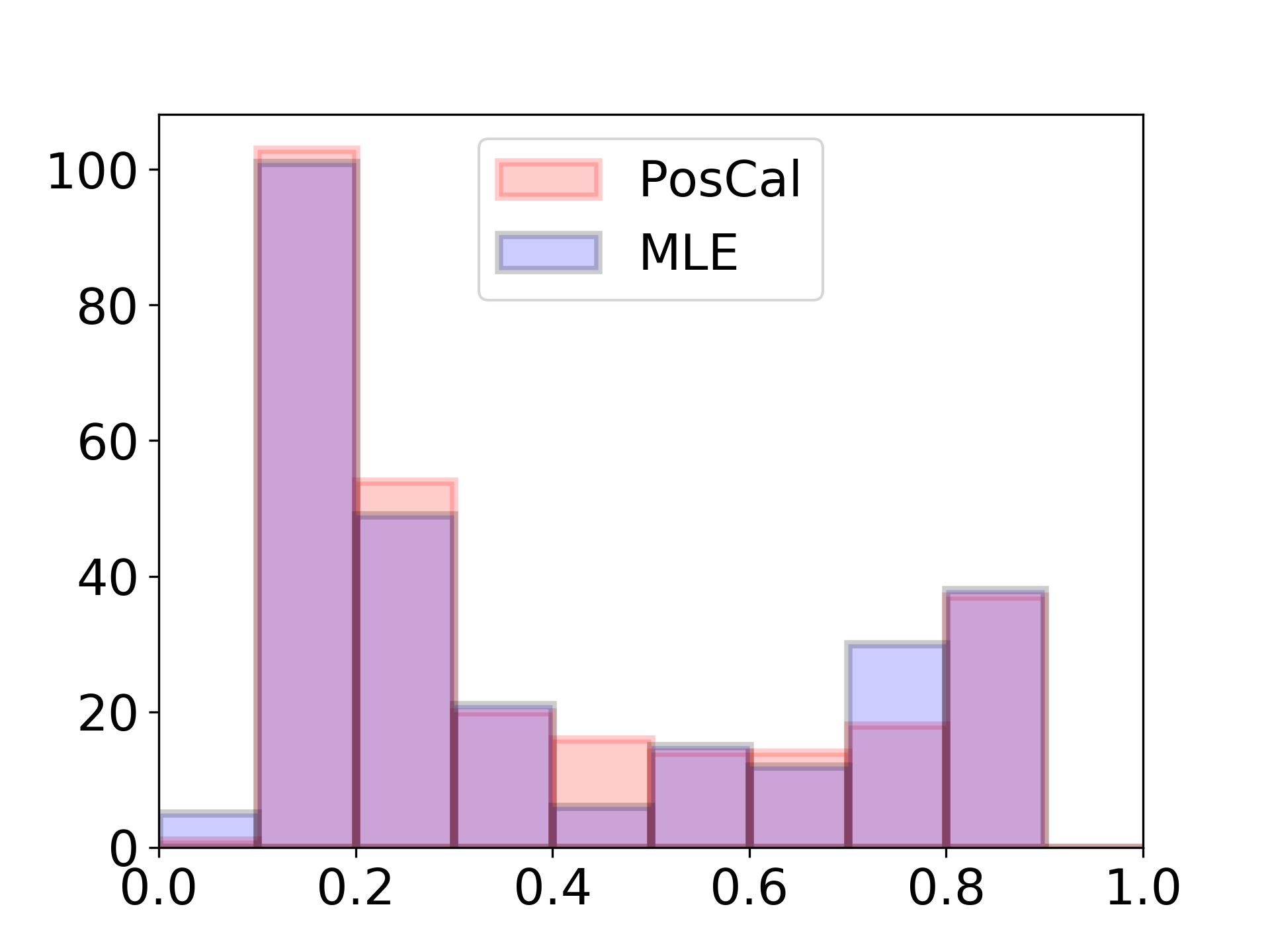}
}
\subfloat[Predictions in SPolite]{
\includegraphics[trim=6mm 6mm 10mm 14mm,clip,width=.48\linewidth]{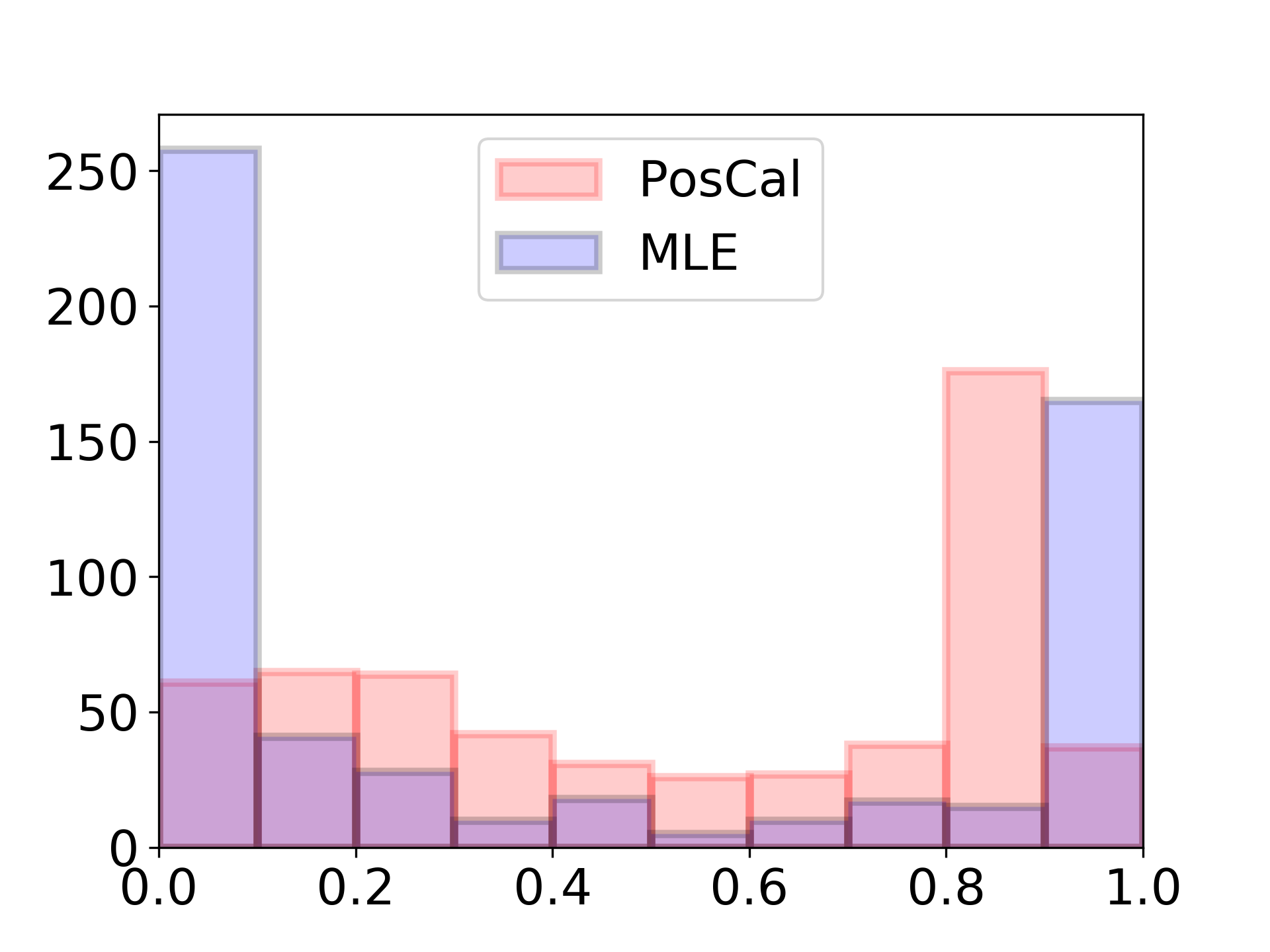}
}
\\
\subfloat[Calibrations in RTE]{
\includegraphics[trim=6mm 6mm 10mm 10mm,clip,width=.48\linewidth]{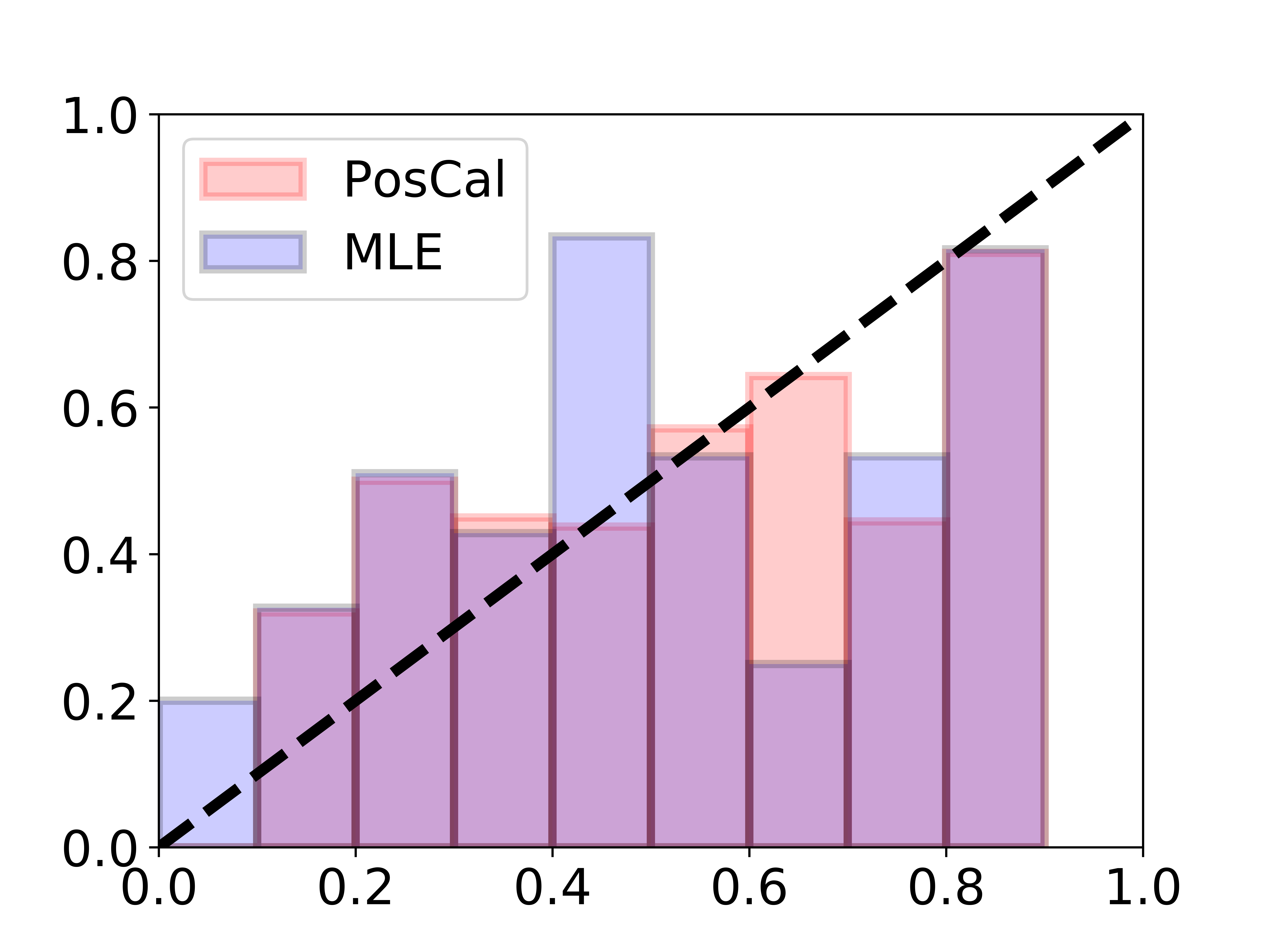}
}
\subfloat[Calibrations in SPolite]{
\includegraphics[trim=6mm 6mm 10mm 10mm,clip,width=.48\linewidth]{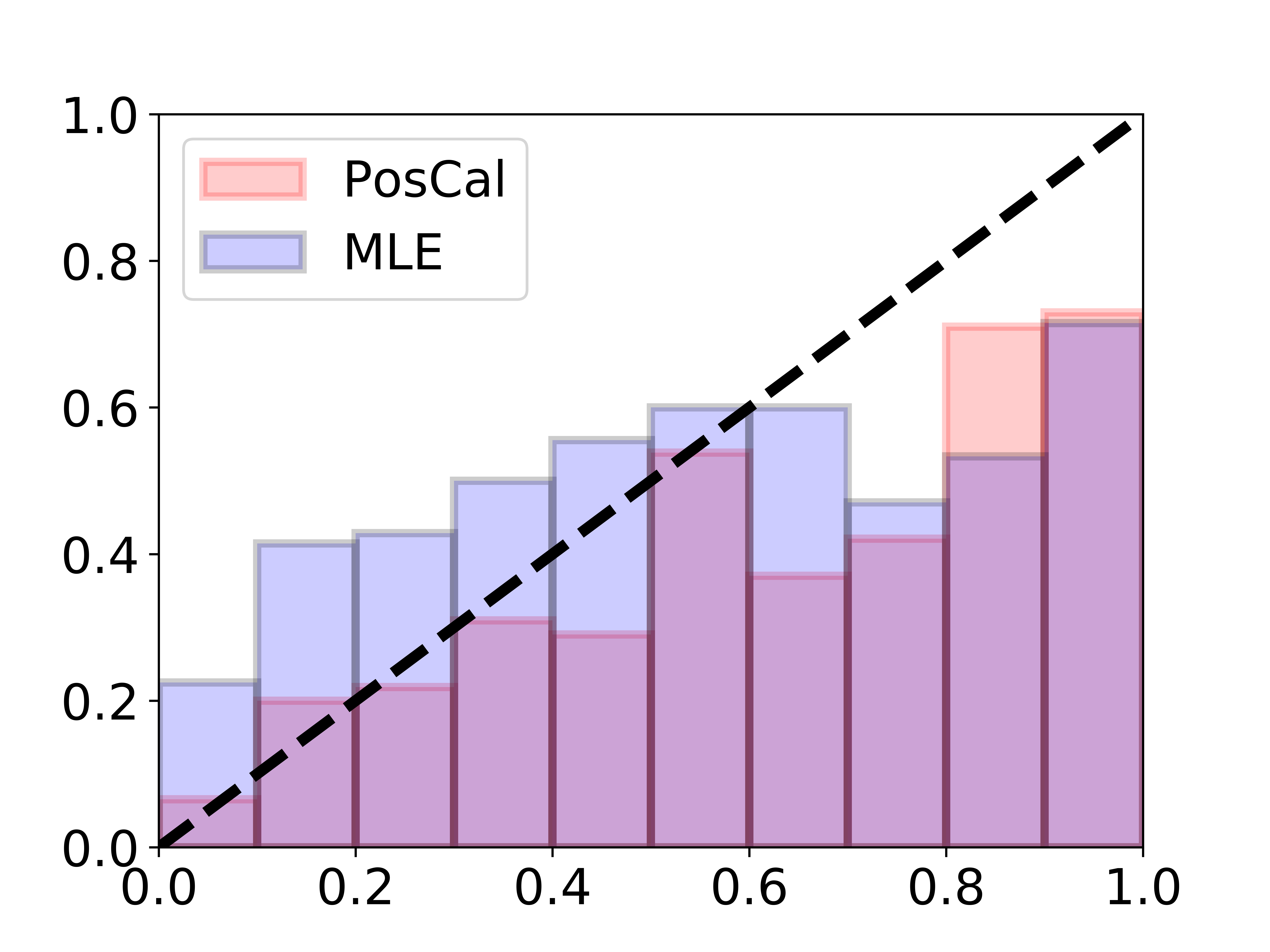}
}
}
\caption{\label{fig:cal_example} 
Histogram of predicted probabilities (top) and their calibration histograms (bottom) between \textbf{MLE} (\colorbox{blue!20}{blue-shaded}) and \textbf{\method} (\colorbox{red!20}{red-shaded}) on RTE in GLUE and SPoliteness in xSLUE. The overlap is \colorbox{purple!40}{purple-shaded}.
X-axis is the predicted posterior, and Y-axis is its frequencies (top) and empirical posterior probabilities (bottom).
The \udensdash{diagonal, linear line} in (c,d) means the expected (or perfectly calibrated) case.
We observe that \colorbox{red!20}{\method} alleviate the posterior probabilities with the small predictions toward \udensdash{the expected calibration}.
Best viewed in color.
}
\end{figure}

\textbf{Analysis.}
We visually check the statistical effect of \method with respect to calibration. Figure~\ref{fig:cal_example} shows how predicted posterior distribution of \textbf{\method} is different from \textbf{MLE}. 
We choose two datasets where \method improves both accuracy and calibration quality compared with the basic MLE: RTE from GLUE and Stanford's politeness dataset from xSLUE.
We then draw two different histograms: a histogram of $\hat{p}$ frequencies (top) and a calibration histogram, $\hat{p}$ versus the empirical posterior probability $q$ (bottom).
Figure~\ref{fig:cal_example}(c,d) show that \method spreads out the extremely predicted posterior probabilities (0 or 1) from MLE to be more well calibrated over different bins.
The well-calibrated posteriors also help correct the skewed predictions in Figure~\ref{fig:cal_example}(a,b).

\begin{table}[t!]
\centering
\small
\begin{tabularx}{\linewidth}{@{} c@{\hskip 0.1cm} c@{\hskip 0.1cm} c@{\hskip 0.1cm} c@{\hskip 0.1cm}  c@{\hskip 0.1cm}c@{\hskip 0.1cm} c@{}}
\toprule
&\scriptsize{\textbf{MLE} $\rightarrow$ \textbf{\method}}
&\textbf{Size } &\textbf{MLE} & \textbf{\method}&
\multicolumn{2}{c}{\textbf{label dist.}} \\
Data &predictions &(\%)& $\texttt{avg}(\hat{p})$&$\texttt{avg}(\hat{p})$ & 0 & 1 \\
\midrule
\parbox[t]{2mm}{\multirow{4}{*}{\rotatebox[origin=c]{90}{{\small\texttt{RTE}}}}} 
&\scriptsize{COR $\rightarrow$ COR}
& 164(59.2)& 79.2& 78.6 & 42.8 & 47.2\\
\cmidrule(r){2-7}
&\scriptsize{\textcolor{red}{COR $\rightarrow$ INCOR}} &3(1.1)& 59.7 & 39.0&0&100\\
&\scriptsize{\textcolor{blue}{INCOR} $\rightarrow$ \textcolor{blue}{COR}} & 9(3.3)& 40.6 & 56.7 &100& 0\\
\cmidrule(r){2-7}
&\scriptsize{INCOR $\rightarrow$ INCOR} & 101(36.4)& 23.6 & 24.9& 27.7&72.3\\
\midrule
\parbox[t]{2mm}{\multirow{4}{*}{\rotatebox[origin=c]{90}{{\small\texttt{SPolite.}}}}} 
&\scriptsize{COR $\rightarrow$ COR} &342(60.3)&95.0&82.6&58.8&41.2\\
\cmidrule(r){2-7}
&\scriptsize{\textcolor{red}{COR $\rightarrow$ INCOR}}&54(9.5)&82.1&26.8&96.3&3.7\\
&\scriptsize{\textcolor{blue}{INCOR} $\rightarrow$ \textcolor{blue}{COR}}  &60(10.6)&16.9&73.9&15.0&85.0\\
\cmidrule(r){2-7}
&\scriptsize{INCOR $\rightarrow$ INCOR} &111(19.6) &9.8 &21.7 &54.0 &46.0\\
\bottomrule
\end{tabularx}
\caption{\label{tab:anal_error} Size of correct (\textbf{COR}) and incorrect (\textbf{INCOR}) prediction labels with their averaged $\hat{p}$(\%) of true labels for \textbf{MLE} and \textbf{\method} on RTE and Stanford's politeness (\textbf{SPolite}) dataset. 
Each has two labels : entail(0) / not entail(1) for RTE, and polite(0) / impolite(1) for SPolite. 
\method improves 2.2\%/1.1\% accuracy than MLE for RTE/SPolite.
}
\end{table}

\begin{table*}[ht!]
\centering
\small
\begin{tabularx}{1.0\textwidth}{@{} P{0.5cm}@{\hskip 0.2cm} p{11.0cm}@{\hskip 0.2cm} P{1.7cm}@{\hskip 0.2cm} P{0.8cm}@{\hskip 0.2cm} P{1.2cm}@{}}
\toprule
\textbf{Data} &\multicolumn{1}{c}{\textbf{Sentence}}&
\textbf{True label} &
\textbf{MLE} $\hat{p}$ & \textbf{\method} $\hat{p}$\\
\midrule
\parbox[t]{2mm}{\multirow{7}{*}{\rotatebox[origin=c]{90}{{\texttt{RTE}}}}}&
\multirow{4}{*}{\parbox{11cm}{\textbf{(S1)} Researchers at the Harvard School of Public Health say that people who drink coffee may be doing a lot more than keeping themselves awake - this kind of consumption apparently also can help reduce the risk of diseases.
\newline
\textbf{(S2)} Coffee drinking has health benefits.}}
&entail	&49.7 &51.3 \\
&&&\multicolumn{2}{c}{\scriptsize\textcolor{blue}{INCOR} $\rightarrow$ \textcolor{blue}{COR}}\\
\\ \\
\cmidrule{2-5}
&\multirow{3}{*}{\parbox{11cm}{\textbf{(S1)} The biggest newspaper in Norway, Verdens Gang, prints a letter to the editor written by Joe Harrington and myself. 
\newline
\textbf{(S2)} Verdens Gang is a Norwegian newspaper.}}
&entail &43.9 &61.9\\
&&&\multicolumn{2}{c}{\scriptsize\textcolor{blue}{INCOR} $\rightarrow$ \textcolor{blue}{COR}}\\
\\
\end{tabularx}
\hspace{\fill}
\begin{tabularx}{1.0\textwidth}{@{} P{0.5cm}@{\hskip 0.2cm} p{11.0cm}@{\hskip 0.2cm} P{1.7cm}@{\hskip 0.2cm} P{0.8cm}@{\hskip 0.2cm} P{1.2cm}@{}}
\toprule
\parbox[t]{2mm}{\multirow{4}{*}{\rotatebox[origin=c]{90}{{\texttt{SPolite.}}}}}
& \multirow{2}{*}{\parbox{11cm}{Not at all clear what you want to do. What is the full expected output?}}
&impolite & 10.5 & 74.9 \\
&&&\multicolumn{2}{c}{\scriptsize\textcolor{blue}{INCOR} $\rightarrow$ \textcolor{blue}{COR}}\\
\cmidrule{2-5}
& \multirow{2}{*}{\parbox{11cm}{Are you sure that it isn't due to the error that the compiler is thrown off, and generating multiple errors due to that one error? Could you give some example of this?}}
&polite&6.9&57.9 \\
&&&\multicolumn{2}{c}{\scriptsize\textcolor{blue}{INCOR} $\rightarrow$ \textcolor{blue}{COR}}\\
\bottomrule
\end{tabularx}
\caption{\label{tab:example} Predicted $\hat{p}$(\%) of true label from \textbf{MLE} and \textbf{\method} with corresponding sentences in RTE and SPolite dataset. 
True label is either entail or not entail for RTE, and polite or impolite for SPolite.
Provided examples are the cases only \method predicts correctly, which correspond to \textcolor{blue}{INCOR} $\rightarrow$ \textcolor{blue}{COR} in table~\ref{tab:anal_error}. 
} 
\end{table*}

To better understand in which case \method helps correct the wrong predictions from MLE, we analyze how prediction $\hat{p}$ is different between MLE and \method in test set.
Table~\ref{tab:anal_error} shows the number of correct/incorrect predictions and its corresponding label distributions grouped by the two models.
For example, COR by MLE and INCOR by \method in the fourth row of Table \ref{tab:anal_error} means that there are three test samples that MLE correctly predicts while \method not. 

We find that in most of cases, \method corrects the wrong predictions from MLE by re-scaling $\hat{p}$ in a certain direction.
In RTE, most inconsistent predictions between MLE and \method have their posterior predictions near to the decision boundary (i.e., 50\% for binary classification) with an averaged predicted probability about 40\%. 
This is mainly because \method does not change the majority of the predictions but helps correct the controversial predictions near to the decision boundary.
\method improves \textcolor{blue}{3.3\%} of accuracy but only sacrifices  \textcolor{red}{1.1\%} by correctly predicting the samples predicted as 'not entailment' by MLE to 'entailment'.

On the other hand, SPolite has more extreme distribution of $\hat{p}$ from MLE than RTE.
We find a fair trade-off between two models (\textcolor{red}{-9.5\%}, \textcolor{blue}{+10.6\%}) but still \method outperforms MLE.


Table~\ref{tab:example} shows examples that only \method predicts correctly, with corresponding $\hat{p}$ of true label from MLE and \method (\textcolor{blue}{INCOR} $\rightarrow$ \textcolor{blue}{COR} cases in Table~\ref{tab:anal_error}). The predicted probability $\hat{p}$ should be greater than $50\%$ if models predict the true label. 

In the first example of RTE dataset, two expressions from S1 and S2 (e.g, ``reduce the risk of disease'' in S1 and ``health benefits'' in S2) make MLE confusing to predict, so $\hat{p}$ of true label becomes slightly less than the borderline probability (e.g., $\hat{p} =  49.7\% < 50\%$), making incorrect prediction.
Another example of RTE shows how the MLE fails to predict the true label since the model cannot learn the connection between the location of newspaper (e.g., ``Norway'') and its name (e.g., ``Verden Gang''). 
In the two cases from SPolite dataset, the level of politeness indicated on phrases (e.g., ``Not at all'' in the first case and ``Could you'' in the second case) is not captured well by MLE, so the model predicts the incorrect label.

From our manual investigation above, we find that statistical knowledge about posterior probability 
helps correct $\hat{p}$ while training \method, so making $\hat{p}$ switch its prediction.
For further analysis, we provide more examples in Appendix C.

\section{Conclusion and Future Directions}\label{sec:conclusion}

We propose a simple yet effective training technique called \method for better posterior calibration.
Our experiments empirically show that \method can improve both the performance of classifiers and the quality of predicted posterior output compared to MLE-based classifiers. 
The theoretical underpinnings of our \method idea are not explored in detail here, but developing formal statistical support for these ideas constitutes interesting future work.
Currently, we fix the bin size at 10 and then estimate $q$ by calculating accuracy of $p$ per bin. 
Estimating $q$ with adaptive binning can be a potential alternative for the fixed binning.


\section*{Acknowledgements}

We thank Matt Gormley and the anonymous reviewers for their helpful comments and discussion.

\bibliographystyle{acl_natbib}
\bibliography{calloss}

\begin{thebibliography}{14}
\expandafter\ifx\csname natexlab\endcsname\relax\def\natexlab#1{#1}\fi

\bibitem[{Card and Smith(2018)}]{card2018importance}
Dallas Card and Noah~A Smith. 2018.
\newblock The importance of calibration for estimating proportions from
  annotations.
\newblock In \emph{Proceedings of the 2018 Conference of the North American
  Chapter of the Association for Computational Linguistics: Human Language
  Technologies, Volume 1 (Long Papers)}, pages 1636--1646.

\bibitem[{Chouldechova(2017)}]{chouldechova2017fair}
Alexandra Chouldechova. 2017.
\newblock Fair prediction with disparate impact: A study of bias in recidivism
  prediction instruments.
\newblock \emph{Big data}, 5(2):153--163.

\bibitem[{Devlin et~al.(2019)Devlin, Chang, Lee, and
  Toutanova}]{devlin2018bert}
Jacob Devlin, Ming-Wei Chang, Kenton Lee, and Kristina Toutanova. 2019.
\newblock \href {https://doi.org/10.18653/v1/N19-1423} {{BERT}: Pre-training of
  deep bidirectional transformers for language understanding}.
\newblock In \emph{Proceedings of the 2019 Conference of the North {A}merican
  Chapter of the Association for Computational Linguistics: Human Language
  Technologies, Volume 1 (Long and Short Papers)}, pages 4171--4186,
  Minneapolis, Minnesota. Association for Computational Linguistics.

\bibitem[{Guo et~al.(2017)Guo, Pleiss, Sun, and
  Weinberger}]{guo2017calibration}
Chuan Guo, Geoff Pleiss, Yu~Sun, and Kilian~Q. Weinberger. 2017.
\newblock \href {http://proceedings.mlr.press/v70/guo17a.html} {On calibration
  of modern neural networks}.
\newblock In \emph{Proceedings of the 34th International Conference on Machine
  Learning}, volume~70 of \emph{Proceedings of Machine Learning Research},
  pages 1321--1330, International Convention Centre, Sydney, Australia. PMLR.

\bibitem[{Kang and Hovy(2019)}]{kang2019xslue}
Dongyeop Kang and Eduard~H. Hovy. 2019.
\newblock xslue: A benchmark and analysis platform for cross-style language
  understanding and evaluation.
\newblock \emph{ArXiv}, abs/1911.03663.

\bibitem[{Kavuluru et~al.(2015)Kavuluru, Rios, and Lu}]{kavuluru2015empirical}
Ramakanth Kavuluru, Anthony Rios, and Yuan Lu. 2015.
\newblock An empirical evaluation of supervised learning approaches in
  assigning diagnosis codes to electronic medical records.
\newblock \emph{Artificial intelligence in medicine}, 65(2):155--166.

\bibitem[{Kull et~al.(2019)Kull, Nieto, K{\"a}ngsepp, Silva~Filho, Song, and
  Flach}]{kull2019beyond}
Meelis Kull, Miquel~Perello Nieto, Markus K{\"a}ngsepp, Telmo Silva~Filho, Hao
  Song, and Peter Flach. 2019.
\newblock Beyond temperature scaling: Obtaining well-calibrated multi-class
  probabilities with dirichlet calibration.
\newblock In \emph{Advances in Neural Information Processing Systems}, pages
  12295--12305.

\bibitem[{Kumar et~al.(2019)Kumar, Liang, and Ma}]{kumar2019verified}
Ananya Kumar, Percy~S Liang, and Tengyu Ma. 2019.
\newblock Verified uncertainty calibration.
\newblock In \emph{Advances in Neural Information Processing Systems}, pages
  3787--3798.

\bibitem[{Nguyen and O{'}Connor(2015)}]{nguyen2015posterior}
Khanh Nguyen and Brendan O{'}Connor. 2015.
\newblock \href {https://doi.org/10.18653/v1/D15-1182} {Posterior calibration
  and exploratory analysis for natural language processing models}.
\newblock In \emph{Proceedings of the 2015 Conference on Empirical Methods in
  Natural Language Processing}, pages 1587--1598, Lisbon, Portugal. Association
  for Computational Linguistics.

\bibitem[{Platt et~al.(1999)}]{platt1999probabilistic}
John Platt et~al. 1999.
\newblock Probabilistic outputs for support vector machines and comparisons to
  regularized likelihood methods.
\newblock \emph{Advances in large margin classifiers}, 10(3):61--74.

\bibitem[{Tibshirani(1996)}]{tibshirani1996regression}
Robert Tibshirani. 1996.
\newblock Regression shrinkage and selection via the lasso.
\newblock \emph{Journal of the Royal Statistical Society: Series B
  (Methodological)}, 58(1):267--288.

\bibitem[{Wang et~al.(2018)Wang, Singh, Michael, Hill, Levy, and
  Bowman}]{wang2018glue}
Alex Wang, Amanpreet Singh, Julian Michael, Felix Hill, Omer Levy, and Samuel
  Bowman. 2018.
\newblock \href {https://doi.org/10.18653/v1/W18-5446} {{GLUE}: A multi-task
  benchmark and analysis platform for natural language understanding}.
\newblock In \emph{Proceedings of the 2018 {EMNLP} Workshop {B}lackbox{NLP}:
  Analyzing and Interpreting Neural Networks for {NLP}}, pages 353--355,
  Brussels, Belgium. Association for Computational Linguistics.

\bibitem[{Zadrozny and Elkan(2001)}]{zadrozny2001obtaining}
Bianca Zadrozny and Charles Elkan. 2001.
\newblock \href {http://dl.acm.org/citation.cfm?id=645530.655658} {Obtaining
  calibrated probability estimates from decision trees and naive bayesian
  classifiers}.
\newblock In \emph{Proceedings of the Eighteenth International Conference on
  Machine Learning}, ICML '01, pages 609--616, San Francisco, CA, USA. Morgan
  Kaufmann Publishers Inc.

\bibitem[{Zadrozny and Elkan(2002)}]{zadrozny2002transforming}
Bianca Zadrozny and Charles Elkan. 2002.
\newblock Transforming classifier scores into accurate multiclass probability
  estimates.
\newblock In \emph{Proceedings of the eighth ACM SIGKDD international
  conference on Knowledge discovery and data mining}, pages 694--699. ACM.

\end{thebibliography}

\renewcommand*\appendixpagename{\Large Appendices}
\clearpage
\begin{appendix}\label{sec:appendix}

\section{Details on Hyper-Parameters}\label{sec:appendix_hyper}

All models are trained with equal hyper-parameters:learning rate 2e-5, and BERT model size BERT$_{BASE}$. 
Also, we set up an early stopping rule for train: we track the validation loss for every 50 steps and then halt to train if current validation loss is bigger than the averaged 10 prior validation losses (i.e., patience 10). 
For \textbf{L1}, we use the regularization weight value 1-e8. For \textbf{\method}, we set up another weight value $\lambda$ for $\Lagr_{Cal}$, and the number of updating empirical probability per epoch ($u$). We tune these two hyper-parameters per each task. For more details, see Table~\ref{tab:app_hyperparams}.  
As a baseline of post-calibration method, we also report ECE with a temperature scaling~\cite{guo2017calibration}, which is current state-of-the-art method.

\begin{table}[h]
\centering
\small
\begin{tabularx}{.8\linewidth}{@{} r @{\hskip 0.2cm} c@{\hskip 0.2cm}c@{\hskip 0.5cm} |r@{\hskip 0.2cm} c@{\hskip 0.2cm}c@{}}
\toprule 
\textbf{xSLUE} & \textbf{$u$} & \textbf{$\lambda$} &  \textbf{GLUE}  & \textbf{$u$} & \textbf{$\lambda$}\\
\midrule
\texttt{GYAFC} & 5&0.6  & \texttt{CoLA} &5&0.2\\
\cmidrule(l{2pt}r{3pt}){1-3}
\texttt{SPolite}& 5&0.6 & \texttt{SST-2} &10&1.0\\
\cmidrule(l{2pt}r{3pt}){1-3} \cmidrule(l{2pt}r{3pt}){4-6}
\texttt{SHumor}& 5&1.0 & \texttt{MRPC} &10&1.0\\
\texttt{SJoke}& 5&1.0 &\texttt{QQP} &10&1.0\\
\cmidrule(l{2pt}r{3pt}){1-3}
\cmidrule(l{2pt}r{3pt}){4-6}
\texttt{SarcGhosh} &5&0.6 &\texttt{MNLI} &2&0.2\\
\texttt{SARC} &5&0.6 &\texttt{MNLI}$_{mm}$ &2&0.2\\
\texttt{SARC}{\_pol}& 5&1.0&\texttt{QNLI} &1&0.6\\
\cmidrule(l{2pt}r{3pt}){1-3}
\texttt{VUA}& 2&1.0&\texttt{RTE} &10&1.0\\
\texttt{TroFi}& 5&1.0&\texttt{WNLI}&2&0.2\\
\cmidrule(l{2pt}r{3pt}){1-3}
\cmidrule(l{2pt}r{3pt}){4-6}
\texttt{CrowdFlower}&5&0.6\\
\texttt{DailyDialog}& 5&1.0\\
\cmidrule(l{2pt}r{3pt}){1-3}
\texttt{HateOffens}& 5&1.0\\
\cmidrule(l{2pt}r{3pt}){1-3}
\texttt{SRomance}& 5&1.0\\
\cmidrule(l{2pt}r{3pt}){1-3}
\texttt{SentiBank}& 5&1.0\\
\cmidrule(l{2pt}r{3pt}){1-3}
\texttt{PASTEL}{\_gender}& 5&1.0\\
\texttt{PASTEL}{\_age}& 5&1.0\\
\texttt{PASTEL}{\_count}& 5&1.0\\
\texttt{PASTEL}{\_polit}& 5&1.0\\
\texttt{PASTEL}{\_educ}& 5&1.0\\
\texttt{PASTEL}{\_ethn}& 5&1.0\\
\bottomrule
\end{tabularx}
\caption{\label{tab:app_hyperparams} 
Hyper-parameters for \method training across tasks : the number of updating empirical probabilities per epoch $u$ and weight value $\lambda$ for $\Lagr_{Cal}$. We tune them using the validation set.
\vspace{-3mm}
} 
\end{table}

\section{Examples When MLE and \method Predicts Different Label}\label{sec:appendix_example}

Table~\ref{tab:example_app} shows some examples in RTE and StanfordPoliteness datasets with their predicted $\hat{p}$ of true label from \textbf{MLE} and \textbf{\method}.

\begin{table*}[ht!]
\centering
\small
\begin{tabularx}{1.0\textwidth}{@{} P{0.5cm}@{\hskip 0.2cm} p{11.0cm}@{\hskip 0.2cm} P{1.7cm}@{\hskip 0.2cm} P{0.8cm}@{\hskip 0.2cm} P{1.2cm}@{}}
\toprule
\textbf{Data} &\multicolumn{1}{c}{\textbf{Sentence}}&
\textbf{True label} &
\textbf{MLE} $\hat{p}$ & \textbf{\method} $\hat{p}$\\
\midrule
\parbox[t]{2mm}{\multirow{32}{*}{\rotatebox[origin=c]{90}{{\texttt{RTE}}}}}&\multirow{5}{*}{\parbox{11.0cm}{
(S1) Charles de Gaulle died in 1970 at the age of eighty. He was thus fifty years old when, as an unknown officer recently promoted to the (temporary) rank of brigadier general, he made his famous broadcast from London rejecting the capitulation of France to the Nazis after the debacle of May-June 1940.	
\newline
(S2) Charles de Gaulle died in 1970.}}	
&entail	&34.9 &58.9 \\
&&&\multicolumn{2}{c}{\scriptsize\textcolor{blue}{INCOR} $\rightarrow$ \textcolor{blue}{COR}}\\
\\ \\ \\
\cmidrule{2-5}
&\multirow{8}{*}{\parbox{11.0cm}{ (S1) Police in the Lower Austrian town of Amstetten have arrested a 73 year old man who is alleged to have kept his daughter, now aged 42, locked in the cellar of his house in Amstetten since 29th August 1984. The man, identified by police as Josef Fritzl, is alleged to have started sexually abusing his daughter, named as Elisabeth Fritzl, when she was eleven years old, and to have subsequently fathered seven children by her. One of the children, one of a set of twins born in 1996, died of neglect shortly after birth and the body was burned by the father.
\newline
(S2) Amstetten is located in Austria.}}	& entail	&45.5	& 57.3 \\
&&&\multicolumn{2}{c}{\scriptsize\textcolor{blue}{INCOR} $\rightarrow$ \textcolor{blue}{COR}}\\
\\ \\ \\ \\ \\ \\
\cmidrule{2-5}
& \multirow{4}{*}{\parbox{11.0cm}{(S1) Swedish massage is used to help relax muscles, increase circulation, remove metabolic waste products, help the recipient obtain a feeling of connectedness, a better awareness of their body and the way they use and position it.
\newline
(S2) Swedish massage loosens tense muscles.}}	& entail &40.1 &56.7 \\
&&&\multicolumn{2}{c}{\scriptsize\textcolor{blue}{INCOR} $\rightarrow$ \textcolor{blue}{COR}}\\
\\ \\
\cmidrule{2-5}
& \multirow{2}{*}{\parbox{11.0cm}{(S1) Blair has sympathy for anyone who has lost their lives in Iraq.
\newline
(S2) Blair is sorry for anyone who has lost their lives in Iraq.}} &entail &31.3 & 50.1 \\
&&&\multicolumn{2}{c}{\scriptsize\textcolor{blue}{INCOR} $\rightarrow$ \textcolor{blue}{COR}}\\
\cmidrule{2-5}
& \multirow{2}{*}{\parbox{11.0cm}{(S1) Capital punishment acts as a deterrent.
\newline
(S2) Capital punishment is a deterrent to crime.}} & entail & 41.6 &64.5 \\
&&&\multicolumn{2}{c}{\scriptsize\textcolor{blue}{INCOR} $\rightarrow$ \textcolor{blue}{COR}}\\
\cmidrule{2-5}
\cmidrule{2-5}
&\multirow{8}{*}{\parbox{11.0cm}{(S1) According to reports, a man protesting the G20 Summit in London, England has died after collapsing at a protester camp. Sky News says the man collapsed on the street inside a camp close to the Bank of England and when found he was still breathing, but efforts by paramedics to rescue him failed and he was pronounced dead at an area hospital. The name of the person and cause of death are not yet known, but several people were injured earlier in the day. It is also reported by Sky News that people threw bottles at him and authorities when they were taking him to a waiting ambulance.
\newline
(S2) Sky News offices are close to the Bank of England.}}
&not entail	& 59.0 &	36.7 \\
&&&\multicolumn{2}{c}{\scriptsize\textcolor{red}{   COR} $\rightarrow$ \textcolor{red}{INCOR}}\\
\\ \\ \\ \\ \\ \\
\cmidrule{2-5}
&\multirow{3}{*}{\parbox{11.0cm}{(S1) The U.S. handed power on June 30 to Iraqâs interim government chosen by the United Nations and Paul Bremer, former governor of Iraq.
\newline
(S2) The United Nations officially transferred power to Iraq. }} &not entail &59.2 &44.9 \\
&&&\multicolumn{2}{c}{\scriptsize\textcolor{red}{   COR} $\rightarrow$ \textcolor{red}{INCOR}}\\
\\
\bottomrule
\end{tabularx}
\begin{tabularx}{1.0\textwidth}{@{} P{0.5cm}@{\hskip 0.2cm} p{11.0cm}@{\hskip 0.2cm} P{1.7cm}@{\hskip 0.2cm} P{0.8cm}@{\hskip 0.2cm} P{1.2cm}@{}}
\midrule
\parbox[t]{2mm}{\multirow{21}{*}{\rotatebox[origin=c]{90}{{\texttt{SPolite.}}}}}
&\multirow{2}{*}{\parbox{11.0cm}{I don't know what page you are talking about, as this is your only edit. Did you perhaps have another account?}}
&\multirow{1}{*}{impolite}	&\multirow{1}{*}{47.3} &\multirow{1}{*}{65.4} \\
&&&\multicolumn{2}{c}{\scriptsize\textcolor{blue}{INCOR} $\rightarrow$ \textcolor{blue}{COR}}\\
\cmidrule{2-5}
&\multirow{2}{*}{\parbox{11.0cm}{Hi.  Not complaining, but why did you remove the category "high schools in california" from this article?}}
&\multirow{1}{*}{impolite} & \multirow{1}{*}{1.2} & \multirow{1}{*}{91.7} \\
&&&\multicolumn{2}{c}{\scriptsize\textcolor{blue}{INCOR} $\rightarrow$ \textcolor{blue}{COR}}\\
\cmidrule{2-5}
& \multirow{2}{*}{\parbox{11.0cm}{Hi, sorry I think I'm missing something here. Why are you adding a red link to the vandalism page?}}
&\multirow{1}{*}{impolite} & \multirow{1}{*}{5.6} & \multirow{1}{*}{61.9} \\
&&&\multicolumn{2}{c}{\scriptsize\textcolor{blue}{INCOR} $\rightarrow$ \textcolor{blue}{COR}}\\
\cmidrule{2-5}
& \multirow{2}{*}{\parbox{11.0cm}{Huh, looks fine to me. Maybe this computer just lies to me to get me to shut up and stop complaining?}}
&\multirow{1}{*}{impolite} & \multirow{1}{*}{3.3} & \multirow{1}{*}{58.1} \\
&&&\multicolumn{2}{c}{\scriptsize\textcolor{blue}{INCOR} $\rightarrow$ \textcolor{blue}{COR}}\\
\cmidrule{2-5}
&\multirow{2}{*}{\parbox{11.0cm}{Can you put an NSLog to make sure it's being called only once? Also, can you show us where you are declaring your int?}} &
\multirow{1}{*}{polite}&\multirow{1}{*}{16.5}&\multirow{1}{*}{76.5}\\
&&&\multicolumn{2}{c}{\scriptsize\textcolor{blue}{INCOR} $\rightarrow$ \textcolor{blue}{COR}}\\
\cmidrule{2-5}
&\multirow{2}{*}{\parbox{11.0cm}{I don't understand the reason for $<$url$>$. Would you please explain it to me?}}
&\multirow{1}{*}{polite}&\multirow{1}{*}{91.5}&\multirow{1}{*}{37.1}\\
&&&\multicolumn{2}{c}{\scriptsize\textcolor{red}{  COR} $\rightarrow$ \textcolor{red}{INCOR}}\\
\cmidrule{2-5}
&\multirow{2}{*}{\parbox{11.0cm}{Another question: Does "Senn" exist in Japanese? If it does, is it possible to render Sennin as Senn-in?}} &\multirow{1}{*}{polite}&\multirow{1}{*}{88.8}&\multirow{1}{*}{45.5}\\
&&&\multicolumn{2}{c}{\scriptsize\textcolor{red}{  COR} $\rightarrow$ \textcolor{red}{INCOR}}\\
\cmidrule{2-5}
& \multirow{3}{*}{\parbox{11.0cm}{Your Chinese is much better than mine. Is it possible in Chinese to omit a conjunction such as u548c or u8ddf and let a string of two successive nouns refer to a set of two objects?}}
&\multirow{1}{*}{polite}&\multirow{1}{*}{94.2}&\multirow{1}{*}{46.4}\\
&&&\multicolumn{2}{c}{\scriptsize\textcolor{red}{  COR} $\rightarrow$ \textcolor{red}{INCOR}}\\
\\
\cmidrule{2-5}
& \multirow{2}{*}{\parbox{11.0cm}{@Smjg, thanks. But why did you also remove the categories I added?}}
&\multirow{1}{*}{impolite}&78.3&45.7\\
&&&\multicolumn{2}{c}{\scriptsize\textcolor{red}{  COR} $\rightarrow$ \textcolor{red}{INCOR}}\\
\cmidrule{2-5}
& \multirow{2}{*}{\parbox{11.0cm}{You can place islands so there is no path between points. What should happen then?}}
&\multirow{1}{*}{impolite}&91.7&35.8\\
&&&\multicolumn{2}{c}{\scriptsize\textcolor{red}{  COR} $\rightarrow$ \textcolor{red}{INCOR}}\\
\bottomrule
\end{tabularx}
\caption{\label{tab:example_app} 
Predicted $\hat{p}$(\%) of true label from \textbf{MLE} and \textbf{\method} with corresponding sentences in RTE (top) and Stanford's politeness (bottom) dataset. 
True label is either entail or not entail for RTE, and polite or impolite for SPolite.
We show the cases where two methods predict the label differently. The case with \textcolor{blue}{INCOR} $\rightarrow$ \textcolor{blue}{COR} means only \method predicts the true label correctly, while the case with \textcolor{red}{COR} $\rightarrow$ \textcolor{red}{INCOR} means only MLE predicts the true label correctly. 
} 
\end{table*}

\end{appendix}

\end{document}